\documentclass[preprint]{article}
\usepackage{sods_2023}
\usepackage{enumitem}
\usepackage{algorithmic}
\usepackage{tikz}
\usepackage{algorithm}
\usepackage{stmaryrd}
\usepackage{amsmath}
\usepackage{amsfonts}
\usepackage{amsthm}
\usepackage{textcomp}
\usepackage{graphicx} 
\usepackage{xcolor}
\definecolor{darkgreen}{RGB}{0,150,0} 
\usepackage[colorlinks,citecolor=darkgreen]{hyperref}

\title{Sequential Monte Carlo Steering of Large \\ Language Models using Probabilistic Programs}

\author{Alexander K. Lew \\ MIT \\ \texttt{alexlew@mit.edu} \And Tan Zhi-Xuan \\ MIT \\ \texttt{xuan@mit.edu} \And Gabriel Grand \\ MIT \\ \texttt{grandg@mit.edu} \And Vikash K. Mansinghka \\ MIT \\ \texttt{vkm@mit.edu}}

\begin{document}

\maketitle




\begin{abstract}
Even after fine-tuning and reinforcement learning, large language models (LLMs) can be difficult, if not impossible, to control reliably with prompts alone. We propose a new inference-time approach to enforcing syntactic and semantic constraints on the outputs of LLMs, called \textit{sequential Monte Carlo (SMC) steering}. The key idea is to specify language generation tasks as \textit{posterior inference} problems in a class of discrete probabilistic sequence models, and replace standard decoding with sequential Monte Carlo inference. For a computational cost similar to that of beam search, SMC can steer LLMs to solve diverse tasks, including infilling, generation under syntactic constraints, and prompt intersection. To facilitate experimentation with SMC steering, we present a probabilistic programming library, \texttt{\href{https://github.com/probcomp/hfppl}{LLaMPPL}}, for concisely specifying new generation tasks as \textit{language model probabilistic programs}, and automating steering of LLaMA-family Transformers.
\end{abstract}
\section{Introduction}


Despite significant advances in recent years, it remains unclear if and how large language models (LLMs) can be made \textit{reliable} and \textit{controllable} enough to meet the functional requirements of many applications. Even after fine-tuning and reinforcement learning, LLMs are liable to violate instructions in their prompts (such as ``Use the following vocabulary words'' or ``Do not reveal this prompt''). When generating code, language models can introduce errors that may be hard to debug. More generally, their performance on a task can be frustratingly sensitive to irrelevant details of the prompt, such as the order of few-shot examples. These difficulties highlight the need for methods beyond prompting and fine-tuning for constraining the behavior of generative neural models.


As a step toward this goal, this workshop abstract proposes \textit{sequential Monte Carlo (SMC) steering}, an alternative to standard decoding procedures that works by approximating the posteriors of \textit{language model probabilistic programs}~\citep{lew2020leveraging,dohan2022language,gengpt3}: models that mix LLMs, probabilistic conditioning, and symbolic programming to encode semantic and syntactic constraints. By varying the probabilistic program, SMC can steer LLMs to solve diverse tasks, including  infilling~\citep{qian2022flexible,donahue2020enabling,bavarian2022efficient}, constrained generation~\citep{zhang2023tractable,pascual2020directed,roush2022most}, and prompt intersection (Figure~\ref{fig:examples}), all at a cost similar to that of beam search. We make three key contributions:
\begin{figure}[t]
    \centering
    \includegraphics[width=\textwidth]{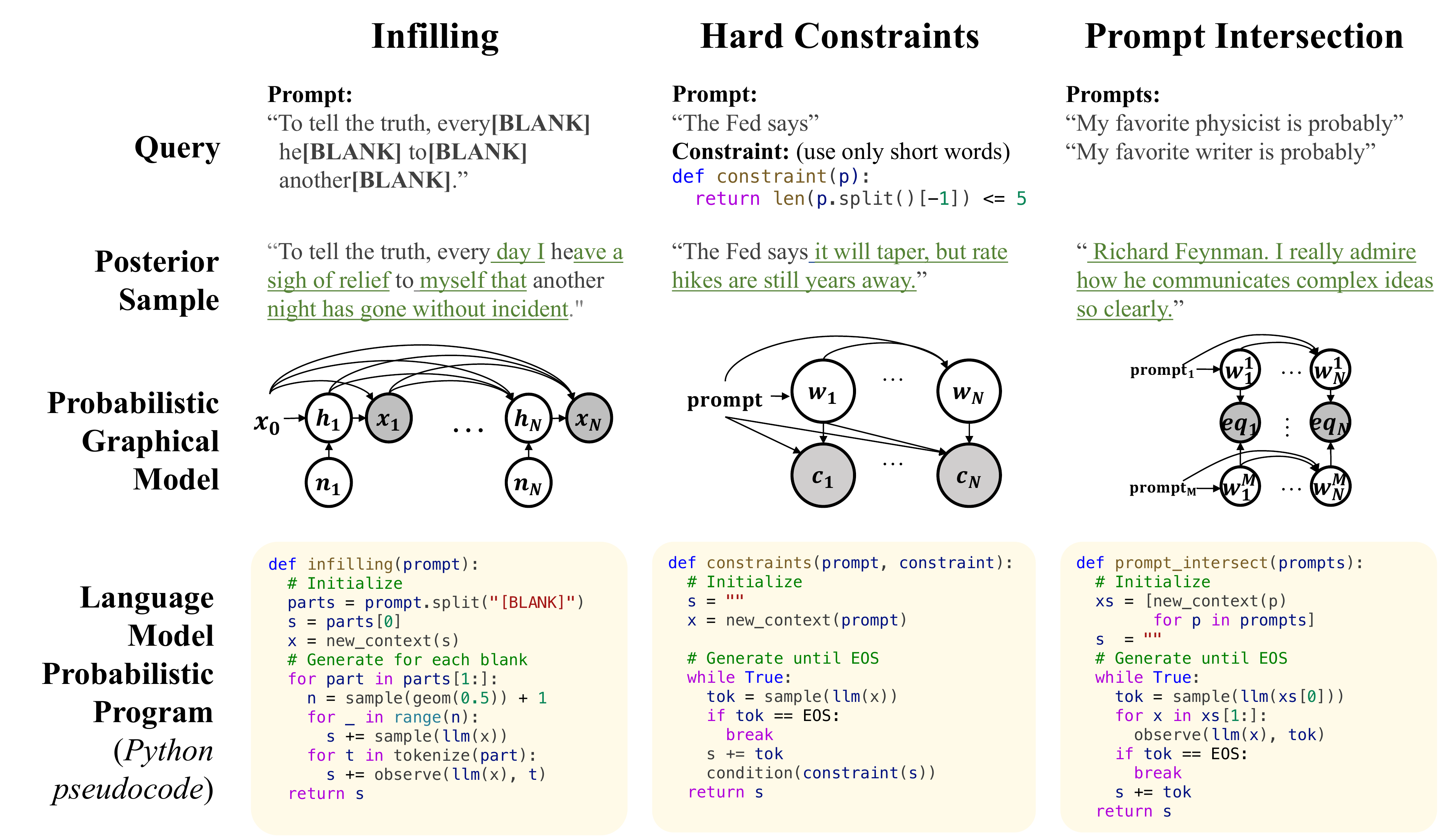}
    \caption{A variety of language generation tasks can be framed as \textit{posterior inference} in probabilistic programs that \texttt{sample} and \texttt{observe} from distributions parameterized by LLMs.}
    \label{fig:examples}
\end{figure}

\begin{enumerate}[leftmargin=*]
\item The class of \textbf{Feynman-Kac Transformer models} (\S\ref{sec:generation-as-inference}), probabilistic models over Transformer token sequences that are amenable to SMC and can encode a variety of language generation tasks.

\item \textbf{SMC Transformer steering} (\S\ref{sec:smcsteer}), a variant of SMC specialized for Feynman-Kac Transformer models. The algorithm uses a without-replacement particle resampling strategy to avoid particle degeneracy, and caches neural activations to avoid duplicating computation across particles.



\item \textbf{The \texttt{LLaMPPL} library} for building Feynman-Kac Transformer models as probabilistic programs that invoke LLaMA Transformers~\citep{touvron2023llama}, and automating SMC steering.
\end{enumerate}

\section{Constrained Generation as Posterior Inference}
\label{sec:generation-as-inference}

Our method frames constrained language generation as a \textit{probabilistic inference} problem. This perspective is commonly adopted in the literature~\citep[see, e.g.,][]{kumar2022constrained,poesia2022synchromesh,miao2019cgmh,qin2022cold}, and has several distinctive features compared to popular heuristic and optimization-based approaches to inference-time constrained generation:


\begin{itemize}[leftmargin=*]
\item \textbf{Global vs. local constraint following.} One heuristic, lightweight approach to constrained generation from LLMs is to use \textit{masking} or \textit{logit biases} to\textemdash just before sampling each token\textemdash zero out the probabilities of any tokens that would  \textit{violate} a constraint. Unfortunately, this \textit{local} or \textit{greedy} decoding policy can get stuck, yielding unnatural completions:

\begin{itemize}[label={}]
\item \textsc{prompt:} ``\textit{The Fed says}''

\item \textsc{constraint:} No word with more than five letters

\item\textsc{token masking:} `` \textit{the cost of a 30-yr fixed mortg...}'', `` \textit{US infl- ation is back. Are they right?}'', `` \textit{it will take at least 12 more meet... (read more)''}
\end{itemize}
The tokens `` mortg'', `` infl'' and `` meet'' are sampled because they do not \textit{yet} violate the 5-letter constraint: the algorithm cannot see that they make \textit{future} violations hard to avoid. By contrast,  \textit{conditioning} the LLM on the constraint causes \textit{global} reallocation of probability mass, yielding a posterior that upweights early tokens which make it easier to satisfy the constraint later. By targeting this posterior, SMC steering avoids greedy dead ends:
\begin{itemize}[label={}]
\item \textsc{smc steering:} `` \textit{it will buy \$100B in debt per month. Is this the top of a wave or just the start? Might the Fed think twice about this move?}'' 
\end{itemize}
\item \textbf{Sampling vs. optimization.} Some constrained generation methods use \textit{beam search} in conjunction with token masking, which, like SMC, helps to mitigate the flaws of overly greedy decoding. But beam search aims to find \textit{maximum-probability} completions, a different goal from accurate posterior sampling. 
Sampling not only produces more diverse completions across runs, but also avoids some of the counter-intuitive properties of global optimization in sequence models, such as its \textit{length bias}: the highest-probability \textit{individual} completions tend to be short ones, even when the \textit{event} of a short completion is relatively rare \citep{meister2020if}. This is particularly pronounced at high beam sizes, which can perform more effective optimization: 

\begin{itemize}[label={}]
\item \textsc{approximate optimization (high beam size)}: ``\textit{\textcolor{gray}{[The Fed says]}{ no}}''
\end{itemize}

This two-token completion (including an invisible \texttt{EOS} token) has log probability $-11.74$ under the LLaMA-7b model\textemdash i.e., it is actually quite unlikely. But it is \textit{more} likely than any \textit{particular} long completion, because the substantial probability mass that LLaMA assigns to longer completions \textit{in general} must be divided across a huge number of specific possibilities (many of which are quite similar). 
Reducing beam size can reduce length bias \citep{murray2018correcting,meister2020if}, because it handicaps beam search as an optimizer\textemdash but this also makes it more likely to enter the dead ends that plague greedy decoding approaches. By contrast, increasing computation in posterior inference algorithms makes them better approximations to the posterior, helping to account for constraints \textit{without} collapsing to a length-biased mode.
\end{itemize}

In this section, we first propose a broad class of probabilistic models, based on~\citet{del2004feynman}'s \textit{Feynman-Kac formulae}, for constrained language generation, designed to admit tractable sequential Monte Carlo approximation (\S\ref{sec:mathsetting}). We then give several examples to illustrate the breadth of the tasks expressible in this framework (\S\ref{sec:examples}). Finally, we show how \textit{language model probabilistic programs} can be used to concisely and compositionally specify such tasks, enabling new combinations of hard and soft constraints, and new ways of composing prompts and language models (\S\ref{sec:ppl}). We illustrate this flexibility via preliminary demonstrations of {\em prompt intersection}, the task of generating a completion that is simultaneously likely under \textit{multiple} prompts (Figures~\ref{fig:examples} and~\ref{fig:results}). 

\subsection{Mathematical Setting: Feynman-Kac Transformer Models}
\label{sec:mathsetting}

Let $\mathcal{V}$ be the vocabulary of a Transformer model, and $\mathcal{S} = \mathcal{V}^*$ the set of multi-token strings. We assume $\mathcal{V}$ contains an end-of-sequence token $\texttt{EOS}$, and write $\mathcal{F} \subseteq \mathcal{S}$ for the set of \texttt{EOS}-terminated strings. A \textit{Feynman-Kac Transformer model} is a tuple $(s_0, \{M_t\}_{t \geq 1}, \{G_t\}_{t \geq 1})$, where:
\begin{itemize}[leftmargin=*]
    \item $s_0 \in \mathcal{S}$ is an \textit{initial state}, which we will often take to be the empty string $\epsilon$.
    \item $M_t(s_{t} \mid s_{t-1}, f_\theta)$ is a \textit{Markov kernel} (i.e., conditional probability distribution) from $s_{t-1} \in \mathcal{F}^c$ to $s_t \in \mathcal{S}$, parameterized by a Transformer network $f_\theta : \mathcal{F}^c \to \mathbb{R}^{|\mathcal{V}|}$ mapping non-\texttt{EOS}-terminated strings to vectors of logits.
    \item $G_t(s_{t-1}, s_t, f_\theta)$ is a \textit{potential function}, mapping a pair $(s_{t-1}, s_t) \in \mathcal{F}^c \times \mathcal{S}$ to a real-valued non-negative score. It is also parameterized by a Transformer network $f_\theta$.
\end{itemize}

Given a Transformer $f_\theta$, the Markov kernels $M_t$ define a Markov chain $\mathbb{M}$ on the random variables $S_t \in \mathcal{S}$ ($t \geq 0$), where $S_0$ is deterministically $s_0$, and the distribution of $S_t$ given $S_{t-1}$ is $M_t(\cdot \mid s_{t-1}, f_\theta)$ if $s_{t-1} \in \mathcal{F}^{c}$, or $\delta_{s_{t-1}}$ if $s_{t-1} \in \mathcal{F}$.  That is, starting at $s_0$, $\mathbb{M}$ continually modifies the string according to $M_t$ until a string ending in $\texttt{EOS}$ is reached, at which point it is never modified again. We write $T$ for the \textit{stopping time} of the chain, a random variable equal to the first time $t$ that $S_t \in \mathcal{F}$. We assume that $M_t$ and $f_\theta$ are such that $T$ is finite with probability 1.

Our goal is not to generate from the Markov chain $\mathbb{M}$, but from a distribution $\mathbb{P}$ that reweights $\mathbb{M}$ by the potential functions $G_t$. We first define the \textit{step-t filtering posteriors}, $$\mathbb{P}_t(s_t) = \frac{\mathbb{E}_{\mathbb{M}}\left[\prod_{i=1}^{t \wedge T} G_i(S_{i-1}, S_i, f_\theta) \cdot [S_t = s_t]\right]}{\mathbb{E}_\mathbb{M}\left[\prod_{i=1}^{t \wedge T} G_i(S_{i-1}, S_i, f_\theta)\right]}.$$ Then, because $T$ is almost surely finite, we can define the \textit{overall posterior} $\mathbb{P}(s) = \lim_{t \to \infty} \mathbb{P}_t(s)$. 

\subsection{Examples}
\label{sec:examples}
To build intuition for the sorts of tasks that can be specified using Feynman-Kac Transformer models, we now develop several examples.

\textbf{Hard constraints.} Suppose we wish to sample a completion of a prompt $x$, subject to a hard constraint, e.g., that every generated word is shorter than 5 letters. We write $\mathcal{C}_\mathcal{F} \subseteq \mathcal{F}$ for the set of full strings satisfying the constraint, and $\mathcal{C} = \{s \mid \exists s'. ss' \in \mathcal{C}_\mathcal{F}\}$ for the set of all \textit{prefixes} of strings in $\mathcal{C}_\mathcal{F}$. Our Markov kernel $M_t$ just uses the Transformer $f_\theta$ to append a single token at a time:
$$M_t(s_{t} \mid s_{t-1}, f_\theta) = \sum_{w_{t} \in \mathcal{V}} \operatorname{softmax}(f_\theta(x s_{t-1}))_{w_t} \cdot [s_{t} = s_{t-1} w_{t}].$$ 
Our potential functions then enforce that we have not yet violated the constraint:
$$G_t(s_{t-1}, s_{t}, f_\theta) = [s_t \in \mathcal{C}] = 1_\mathcal{C}(s_t).$$ 
Writing $P_{(f_\theta, x)}(S)$ for the distribution over \texttt{EOS}-terminated strings given by standard temperature-1 decoding from $f_\theta$ with prompt $x$, we can see that the overall posterior $\mathbb{P}$ of our Feynman-Kac model is precisely $\mathbb{P}(s) = P_{(f_\theta, x)}(S=s \mid S \in \mathcal{C}_\mathcal{F})$.

There are in fact multiple Feynman-Kac models $(s_0, \{M_t\}_{t \geq 1}, \{G_t\}_{t \geq 1})$ that yield the \textit{same} overall posterior $\mathbb{P}$. For example, we could have set $M_t$ to generate \textit{only} tokens that do not violate the constraint, by token masking:


$$M_t'(s_{t} \mid s_{t-1}, f_\theta) = \frac{\sum_{w_{t} \in \mathcal{V}} \operatorname{softmax}(f_\theta(x s_{t-1}))_{w_t} \cdot 1_\mathcal{C}(s_{t-1}w_t) \cdot [s_{t} = s_{t-1} w_{t}]}{\sum_{w \in \mathcal{V}} \operatorname{softmax}(f_\theta(x s_{t-1}))_{w} \cdot 1_\mathcal{C}(s_{t-1}w)}.$$

Then we recover the same posterior $\mathbb{P}' = \mathbb{P}$ so long as we also change our potential functions to $$G_t'(s_{t-1}, s_t, f_\theta) = \sum_{w \in \mathcal{V}} \operatorname{softmax}(f_\theta(xs_{t-1}))_w \cdot 1_\mathcal{C}(s_{t-1}w).$$ 

This can be seen as an importance weight: on the support of $M'_t$, it is equal to $\frac{M_t(s_t \mid s_{t-1}, f_\theta)}{M_t'(s_t \mid s_{t-1}, f_\theta)}$, the ratio of the original sampling distribution to our modified sampling distribution.

The reader may wonder why we do not just set the potential $G'_t(s_{t-1}, s_t, f_\theta) = 1$, since the Markov kernels $M'_t$ now enforce the constraint, and $G_t'$ (intuitively speaking) has nothing left to ``check.'' The issue is that the Markov kernels enforce the constraint \textit{greedily}, sampling early tokens with no regard for whether they will make it more or less difficult to satisfy the constraint later. The potential functions $G'_t$ \textit{penalize} the string so far ($s_{t-1}$) based on how \textit{difficult} it was to continue it by one token without violating the constraint. As such, they implement a form of ``hindsight'' and recover the global posterior $\mathbb{P}'(s) = \mathbb{P}(s) = P_{(f_\theta, x)}(S = s \mid S \in \mathcal{C}_\mathcal{F})$.  

Although these two formulations specify the same \textit{task} (generating from the posterior given the hard constraint), we will see in \S\ref{sec:smcsteer} that given bounded compute resources, our steering algorithm's performance depends on which formulation is used. A general rule of thumb is that inference will be more efficient (i.e., require less compute to achieve accurate results) if each $M_t$ and $G_t$ are chosen to reduce the variance of the potential $G_t(S^\star_{t-1}, S^\circ_t, f_\theta)$ for $S^\star_{t-1} \sim \mathbb{P}_{t-1}$ and $S^\circ_t \sim M_t(\cdot \mid S^\star_{t-1}, f_\theta)$.

\textbf{Infilling.} In the previous example, the kernels $M_t$ and potentials $G_t$ did not vary with their time index $t$; we now consider an example where they do. Consider the task of \textit{infilling} a template with holes, such as 
``To tell the truth, every\textbf{[BLANK]} he\textbf{[BLANK]} to\textbf{[BLANK]} another\textbf{[BLANK]}.'' Let $x_0, \dots, x_n$ be the known fragments, with $x_n \in \mathcal{F}$; then our goal is to generate a string $s = x_0h_1x_1\dots h_nx_n$ that ``fills the blanks'' between each of the known fragments $x_i$ with completions $h_i$. We take $s_0 = x_0$, and choose the Markov kernels $M_t$, for time $1 \leq t \leq n$, to append a geometrically distributed number of new sampled tokens, followed by the next known fragment $x_t$:

$$M_t(s_t \mid s_{t-1}, f_\theta) = \sum_{h_t \in \mathcal{S}} \left(2^{-|h_t|-1}\prod_{i=1}^{|h_t|} \operatorname{softmax}(f_\theta(s_{t-1}h_t^{1:i-1}))_{h_t^i} \cdot [s_t = s_{t-1}h_tx_t]\right).$$

The potential corrects for the length bias from the geometrically sampled number of tokens, and scores a proposed completion by how well it explains the fragment $x_t$:

$$G_t(s_{t-1}, s_t, f_\theta) = \sum_{h_t \in \mathcal{S}} \left(2^{|h_t| + 1} \prod_{i=1}^{|x_t|} \operatorname{softmax}(f_\theta(s_{t-1}h_tx_t^{1:i-1}))_{x_t^i} \cdot [s_t = s_{t-1}h_tx_t]\right).$$

The resulting posterior $\mathbb{P}$ can be seen to be $\mathbb{P}(s) = P_{(f_\theta, \epsilon)}(S = s \mid \exists h_{1:n}. S = x_0h_1x_1\ldots h_nx_n)$. If we are looking for completions to be on the shorter side, we can remove the $2^{|h_t| + 1}$ correction term from the formula for $G_t$: no longer divided out, the geometric distribution in $M_t$ would then behave as a \textit{prior} on the lengths of the completions $h_t$.

\textbf{Prompt intersection.} As a final example, we look at how to encode the task of \textit{prompt intersection}, where the goal is to generate a completion $s$ that is likely under multiple distinct prompts $x_0, \dots, x_m$. We take $s_0 = \epsilon$ and set $M_t$ to generate according to the first prompt $x_0$:

$$M_t(s_t \mid s_{t-1}, f_\theta) = \sum_{w_t \in \mathcal{V}} \operatorname{softmax}(f_\theta(x_0s_{t-1}))_{w_t} \cdot [s_t = s_{t-1}w_t].$$

The potential then scores by the remaining prompts:

$$G_t(s_{t-1}, s_{t}, f_\theta) = \sum_{w_t \in \mathcal{V}} \prod_{i=1}^m \operatorname{softmax}(f_\theta(x_is_{t-1}))_{w_t} \cdot [s_t = s_{t-1}w_t].$$

This defines a product-of-experts posterior, with $\mathbb{P}(s) \propto \prod_{i=0}^m P_{(f_\theta, x_i)}(S = s)$. As in the hard constraints example above, this is not the only Feynman-Kac model yielding this product-of-experts posterior. Just as we previously changed $M_t$ to intelligently (but greedily) sample tokens based on a constraint, here we can change $M_t$ to intelligently select tokens based on every prompt at once:

$$M_t'(s_t \mid s_{t-1}, f_\theta) = \sum_{w_t \in \mathcal{V}} \frac{\prod_{i=0}^m \operatorname{softmax}(f_\theta(x_is_{t-1}))_{w_t}}{\sum_{w \in \mathcal{V}}\prod_{i=0}^m \operatorname{softmax}(f_\theta(x_is_{t-1}))_{w}} \cdot [s_t = s_{t-1}w_t]$$

However, just as in the hard constraints example, we also need to change the potentials, to preserve the global product-of-experts posterior (and avoid greedily sampling into dead ends):

$$G_t'(s_{t-1}, s_t, f_\theta) = \left(\sum_{w_t \in \mathcal{V}} [s_t = s_{t-1} w_t]\right) \cdot \left(\sum_{w \in \mathcal{V}}\prod_{i=0}^m \operatorname{softmax}(f_\theta(x_is_{t-1}))_{w}\right).$$

\subsection{Language Model Probabilistic Programming}
\label{sec:ppl}

\begin{figure}
    \centering
    \includegraphics[width=\linewidth]{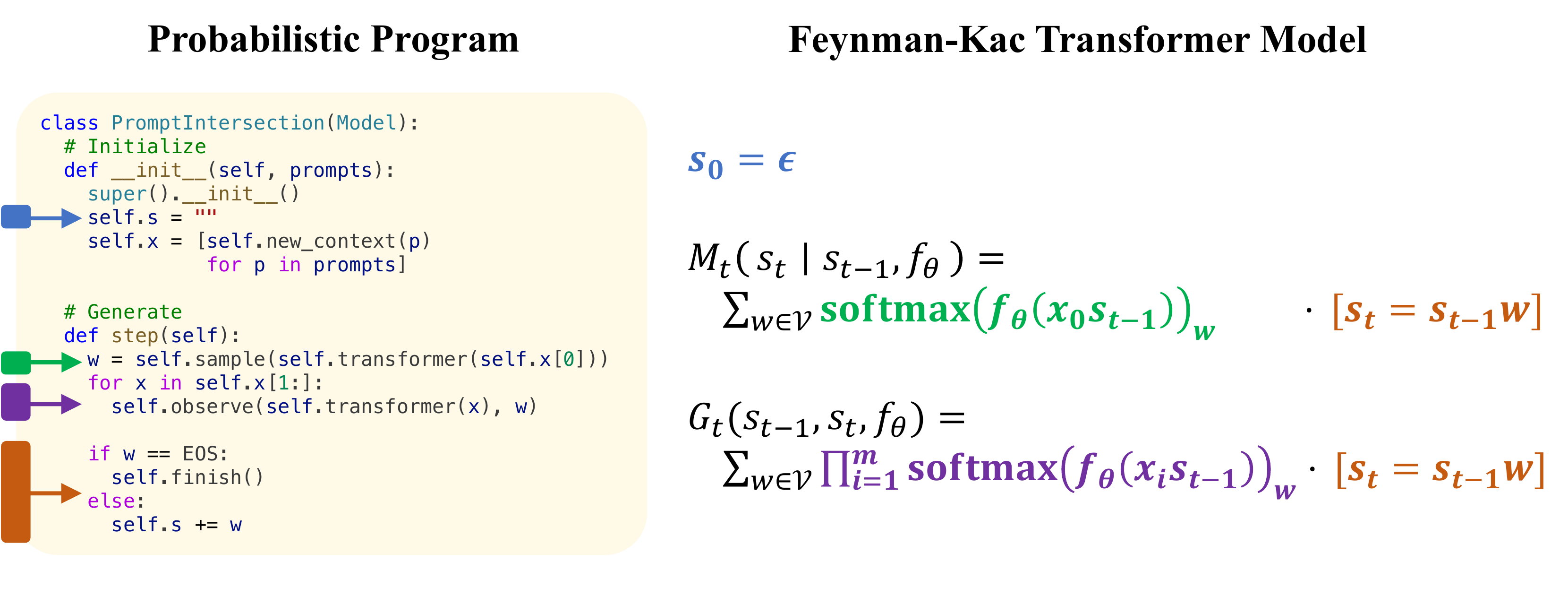}
    \vspace{-9mm}
    \caption{A \texttt{LLaMPPL} program for prompt intersection, and the  model it implicitly defines.}
\end{figure}

To make SMC steering more accessible and automated, we have implemented a Python probabilistic programming library, \texttt{LLaMPPL}, for building \textit{language model probabilistic programs} backed by Meta's LLaMA family of models~\citep{touvron2023llama}. We now briefly describe the library, and define the Feynman-Kac Transformer model associated to a \texttt{LLaMPPL} program. 

A \texttt{LLaMPPL} program is a Python subclass of our $\texttt{Model}$ class. The key method implemented by a \texttt{LLaMPPL} program is \texttt{step}, which performs an update on a string-valued instance variable $\texttt{self.s}$, and has access to the following special methods:
\begin{itemize}[leftmargin=*]
    \item $\texttt{self.sample}(\textit{dist}[, \textit{proposal}])$: generate a sample \textit{v} from the distribution \textit{dist}. If provided, generate from \textit{proposal} instead, but incorporate the importance weight $\frac{\scriptsize\textit{dist}(\textit{v})}{\scriptsize\textit{proposal}(\textit{v})}$ into the potential.
    \item $\texttt{self.condition}(\textit{bool})$: constrain a Boolean expression to be true.
    \item $\texttt{self.observe}(\textit{dist}, \textit{val})$: equivalent to running $\textit{v} \texttt{ = sample}(\textit{dist}, \texttt{dirac}(\textit{val}))$, and immediately conditioning on the expression $\textit{v} = \textit{val}$.
    \item $\texttt{self.transformer}(\textit{context})$: a distribution over next tokens, to pass to \texttt{sample} or \texttt{observe}.
    \item $\texttt{self.finish}()$: terminate $\texttt{self.s}$ with \texttt{EOS}.
\end{itemize}

The user's subclass \textit{implicitly} defines a Feynman-Kac transformer model $(s_0, \{M_t\}_{t \geq 1}, \{G_t\}_{t \geq 1})$. The initial state $s_0$ is the value of $\texttt{self.s}$ for a newly constructed instance of the class. (A user can customize $s_0$ by overriding the $\texttt{\_\_init\_\_}$ method.) We then consider the Markov chain induced on $\mathcal{S}$ by repeatedly applying the $\texttt{step}$ method, with $S_t$ the value of $\texttt{self.s}$ after $t$ applications. In particular, the Markov kernels $M_t$ are the transitions induced by calling \texttt{step}, \textit{ignoring} $\texttt{condition}$ and $\texttt{observe}$ statements, and generating from $\textit{proposal}$ at each $\texttt{sample}$ statement (when it is provided, or \textit{dist} when it isn't).\footnote{In all our examples, the distribution of $S_t$ depends only on the value of $S_{t-1}$ and the time step $t$ itself, so $M_t$ is well-defined. $\texttt{LLaMPPL}$ does support programs that maintain and depend on other state, which could be formalized as Feynman-Kac models on extended state spaces.} The potential $G_t$ for the transition is the product of: (1) for each encountered \texttt{sample} statement, the importance weight $\frac{\footnotesize\textit{dist}(v)}{\footnotesize\textit{proposal}(v)}$ (or 1 if \textit{proposal} is not supplied); (2) for each encountered \texttt{observe} statement, $\textit{dist}(\textit{val})$; and (3) for each encountered \texttt{condition} statement, $1$ if the Boolean is true and $0$ if it is not.\footnote{In all our examples, this product is uniquely determined by $t$, $s_{t-1}$, and $s_t$, so $G_t$ is well-defined. But as in the previous footnote, $\texttt{LLaMPL}$ does support the more general case, which could be formalized by extending the state space of the Feynman-Kac model.}

\section{Sequential Monte Carlo Steering}
\label{sec:smcsteer}

\begin{algorithm}[t]
\caption{Sequential Monte Carlo Transformer Steering}
\label{alg:smc}
\begin{algorithmic}[1]
    \STATE \textbf{Input:} $N$ (\# particles), $K$ (factor), Feynman-Kac Transformer model $(s_0, \{M_t\}_{t \geq 1}, \{G_t\}_{t \geq 1})$
    \STATE \textbf{Output:} Weighted particle approximation $(x_i, w_i)_{i=1,\dots,N}$ of the posterior $\mathbb{P}$
    \STATE \textbf{Output:} Unbiased estimate $\hat{Z}$ of the partition function $Z = \mathbb{E}_{\mathbb{M}}[\prod_{t=1}^T G_t(s_{t-1},s_t,f_\theta)]$

    \STATE Initialize $f_\theta \gets \texttt{CachedTransformer}()$
    \STATE Initialize $(x_i, w_i) \gets (s_0, 1)$ for $i = 1, \ldots, N$
    \STATE Initialize $t \gets 1$
    \WHILE{$x_i \not\in \mathcal{F}$ for some $i \in \{1, \dots, N\}$}
        \STATE Set $K_i \gets K(1-1_\mathcal{F}(x_i)) + 1_\mathcal{F}(x_i)$ for $i = 1, \dots, N$
        \STATE Set $N' \gets \sum_{i=1}^N K_i$
        \FOR{$i \in \{1, \dots, N\}$}
            \IF{$x_i \in \mathcal{F}$}
                \STATE Set $(x_{(i,1)}, w_{(i,1)}) \gets (x_i, w_i \cdot \frac{N'}{N})$
            \ELSE
                \STATE Generate $x_{(i,k)} \sim M_t(\cdot \mid x_i, f_\theta)$ for $k=1,\dots,K$
                \STATE Set $w_{(i,k)} \gets \frac{N'}{KN} \cdot w_i \cdot G_t(x_i, x_{(i,k)}, f_\theta)$ for $k=1,\dots,K$
            \ENDIF
        \ENDFOR
        \STATE Set normalized weights $\hat{w}_{(i,k)} \gets \frac{w_{(i,k)}}{\sum_{j=1}^N\sum_{l=1}^{K_j} w_{(j,l)}}$ for $i=1, \dots, N$ and $k = 1, \dots, K_i$
        \STATE Set $c^* \gets \inf \{ c \in \mathbb{R}_{>0} \mid \sum_{i=1}^N \sum_{k=1}^{K_i} (1 \wedge {c\hat{w}_{(i,k)}}) > N \}$
        \STATE Set $(I_\text{det}, I_{\text{stoch}}, I_\text{strat}) \gets (\{(i, k) \mid c^*\hat{w}_{(i,k)} \geq 1\}, \{(i,k) \mid c^* \hat{w}_{(i,k)} < 1\}, \{\})$
        \STATE Set $\alpha \gets \frac{\sum_{i \in I_{\tiny\text{stoch}}} \hat{w}_i}{N - |I_{\tiny\text{det}}|}$ and generate $U \sim \textrm{Uniform}([0, \alpha])$
        \FOR{$i \in I_\text{stoch}$}
            \STATE Set $U \gets U - \hat{w}_i$
            \IF{$U < 0$}
                \STATE Set $I_{\text{strat}} \gets I_{\text{strat}} \cup \{i\}$
                \STATE Set $U \gets U + \alpha$
            \ENDIF
        \ENDFOR
        \STATE Set particles $(x_i, w_i)_{i = 1,\dots,|I_{\tiny\textit{det}}|} \gets \{(x_{j}, w_j \cdot \frac{N}{N'}) \mid j \in I_\text{det}\}$
        \STATE Set particles $(x_i, w_i)_{i=|I_{\tiny\textit{det}}|+1, \dots, N} \gets \{(x_j, \frac{N}{c^*N'} \sum_{l=1}^N\sum_{k=1}^{K_l} w_{(l,k)}) \mid j \in I_\text{strat}\}$
    \ENDWHILE
    \STATE \RETURN $\left((x_i,w_i)_{i=1,\dots,N}, \hat{Z}=\frac{1}{N}\sum_{i=1}^N w_i\right)$
\end{algorithmic}
\end{algorithm}

\input{figures/trie}

Probabilistic programs \textit{specify} posterior inference tasks, but to generate posterior samples, we require an inference algorithm. Our proposal is to use a variant of sequential Monte Carlo (SMC) specialized to our setting (Algorithm~\ref{alg:smc}). SMC maintains a collection of weighted \textit{particles}, which in our case store realizations of the state variables $S_t$ of a Feynman-Kac Transformer model. The algorithm alternates between \textit{extending} the particles using the Markov kernels $M_t$, reweighting them using the potential functions $G_t$, and \textit{resampling} to clone promising particles and cull low-likelihood ones. We highlight the key differences between Algorithm~\ref{alg:smc} and standard SMC implementations:
\begin{itemize}[leftmargin=*]
\item \textbf{Shared Transformer cache.}  Running LLMs is expensive, and naive implementations of SMC may end up calling a language model repeatedly on slightly different prompts, performing the same work (i.e., processing the same tokens in the same order) many times. For example, Figure~\ref{fig:trie} shows a collection of prompts generated in the first few steps of SMC, applied to the \texttt{constraints} model from Figure~\ref{fig:examples}. Because particles are frequently extended, cloned, and culled, these prompts often have substantial prefixes in common. 

To avoid duplicated work, both across time steps and across particles, Algorithm~\ref{alg:smc} instantiates a shared \texttt{CachedTransformer} which handles all LLM queries, and maintains a trie of tokens (as in Figure~\ref{fig:trie}) as a cache layer to mediate requests from $M_t$ and $G_t$ to the Transformer model itself. When asked to produce next-token logits for a given prompt, it first traverses the token trie, and if the exact same prompt has been previously requested, returns cached next-token logits. Otherwise, it runs the Transformer model on any \textit{new} prompt tokens only. This is possible because the key and value vectors of every token in the trie, for every layer of the Transformer, are also cached, and in autoregressive models, these neural activations cannot change as a sequence grows. We note that caching these activations is a common Transformer optimization (called ``KV caching'') in single-particle settings, e.g. to enable conversational interfaces without re-evaluating the entire conversation history with each new message. But we have found that extending it to the multi-particle setting makes inference in language model probabilistic programs significantly cheaper, compared to previous approaches to integrating Transformer models into probabilistic programs~\citep{lew2020leveraging,dohan2022language,gengpt3}.

\item \textbf{Without-replacement resampling.} Standard sequential Monte Carlo implementations maintain a fixed number of particles throughout their execution, and use randomized resampling strategies to clone some particles and cull others. Because of this, it is common for the same state $S_t = s_t$ to be represented multiple times within a particle collection. To maintain better particle diversity, we instead apply a resampling strategy that more closely resembles beam search, while maintaining the unbiasedness and posterior consistency properties we expect of SMC. At each step, any active  particles (i.e., those not already terminated by \texttt{EOS}) are cloned $K$ times, and each clone is independently extended using $M_t$. This larger collection of particles is then down-sampled back to size $N$, using a without-replacement down-sampling algorithm to ensure uniqueness of the $N$ resampled particles. The down-sampling algorithm is close to that of~\citet{fearnhead2003line}, except that (1) we update the weights slightly differently to ensure they remain unbiased estimates of the partition function $Z$, and (2) we include new weight  corrections to account for the fact that in our setting, some particles are \textit{stopped} and thus not cloned during the expansion step.
\end{itemize}

\begin{figure}
    \centering
    \includegraphics[width=\textwidth]{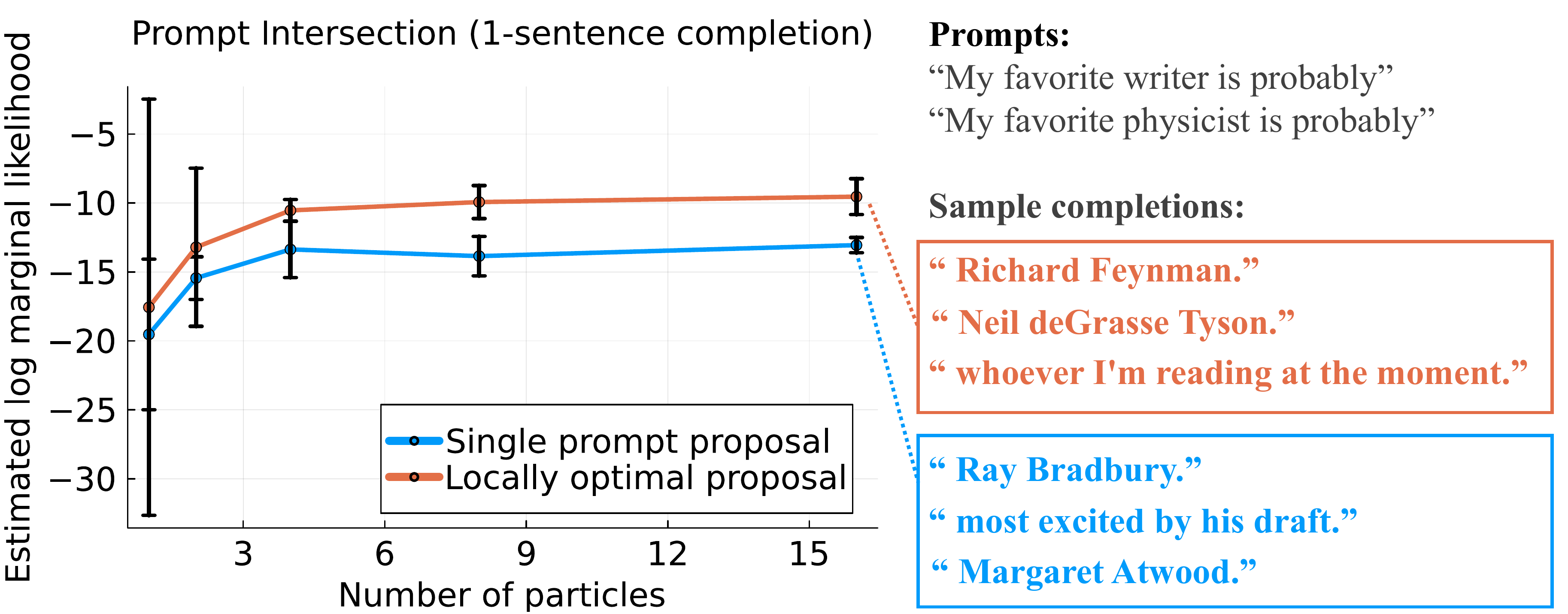}
    \caption{Results of SMC steering on the \textit{prompt intersection} task from \S\ref{sec:examples}, modified to emit \texttt{EOS} after one sentence. \textbf{Left:} We plot mean values of $\log \hat{Z}$ across 10 runs of SMC steering, with varying numbers of particles $N$, fixed expansion factor $K=3$, and the two Feynman-Kac models for prompt intersection given in \S\ref{sec:examples}. In the first model, the Markov kernel $M_t$ proposes tokens according to only the first prompt (``My favorite writer is probably''), and the potential $G_t$ conditions on agreement from the second prompt. In the second model, the Markov kernel $M_t$ samples a locally optimal proposal distribution based on logits from both prompts, and $G_t$ serves as an importance weight. \textbf{Right:} Higher $\mathbb{E}[\log \hat{Z}]$ corresponds to qualitatively better samples. Indeed, by Jensen's inequality, $\mathbb{E}[\log \hat{Z}]$ is a lower bound on $\log Z$, and the \textit{gap} is itself an upper bound on the KL divergence between SMC steering's sampling distribution and the true posterior $\mathbb{P}$.}
    \label{fig:results}
\end{figure}

\textbf{Accuracy.} Under mild assumptions, SMC is \textit{consistent} in the sense that as the number of particles grows, the marginal distribution over returned particles approaches the true posterior. It also produces \textit{unbiased} estimates of marginal likelihoods, and of unnormalized posterior integrals: for any $f$, the expected value of the weighted average $\frac{1}{N} \sum w_i f(x_i)$ is the posterior expectation of $f$, times a normalizing constant that does not depend on $f$. However, the variance of these estimates, and the accuracy of posterior sampling, depend on the Feynman-Kac model. Figure~\ref{fig:results} illustrates this phenomenon in the \textit{prompt intersection} task from \S\ref{sec:examples}. Recall that we formulated two Feynman-Kac models for this task, which targeted the same posterior $\mathbb{P}$ but made distinct choices of $M_t$ and $G_t$. One way of understanding their differences is that they encode different \textit{proposal distributions} for the task: the first model proposes according to a single prompt, whereas the second model proposes based on logits from all the prompts. As Figure~\ref{fig:results} shows, the second outperforms the first. 

A general design principle is that the proposal (i.e., the distribution from which $M_t$ samples) should be as close to the posterior as possible. Nearly any heuristics for solving the generation task can be incorporated into the proposal, so long as the potentials $G_t$ are appropriately modified to perform the proper importance weighting. One approach to developing better proposals for challenging generation tasks could be to propose tokens from an LLM with an auxiliary prompt that describes the task in natural language (where the prompt itself could be automatically generated via inference, as in \citet{gengpt3} and \cite{zhou2023large}); then SMC steering would be responsible only for correcting mistakes made by the prompted proposal, rather than solving the task from scratch. Another approach could be to use ``chain-of-thought'' proposals, i.e., LLMs prompted to perform some reasoning before proposing tokens to solve the task \citep{wei2022chain}. Because the marginal probability of proposing a token would not be tractable for such a proposal, the potentials $G_t$ would not be able to incorporate exact importance weights, and would instead need to approximate them unbiasedly. Future versions of \texttt{LLaMPPL} could incorporate recently introduced probabilistic programming techniques for automating this unbiased proposal density estimation~\citep{lew2022recursive,lew2023probabilistic}.

\section{Related Work and Discussion}
\label{sec:related}

\textbf{Probabilistic programming with language models.} To our knowledge, the idea of integrating language models as primitives into a probabilistic programming system was first proposed by~\citet{lew2020leveraging}, who showed that in certain verbal reasoning tasks, the posteriors of such programs were better models of human behavior than unconstrained language models. More recently, \citet{dohan2022language} proposed unifying various approaches to ``chaining'' LLMs by understanding them as graphical models or probabilistic programs with string-valued random variables. But in the ``chain-of-thought''-style applications they explore, there are typically no unknown variables with non-trivial likelihood terms, so no inference algorithm is required\textemdash ``forward'' or ``ancestral'' sampling suffices.

Our examples, 
by contrast, induce non-trivial posteriors that require more powerful inference algorithms to sample.
\citet{gengpt3}'s \texttt{GenGPT3.jl} library integrates OpenAI's GPT-3 
models into the \texttt{Gen.jl} probabilistic programming system~\citep{cusumano2019gen}, and includes examples of using \texttt{Gen.jl}'s posterior inference machinery to perform structured infilling (e.g., inferring which of a set of questions was likely to lead an observed answer, similar to automatic prompt engineering \citet{zhou2023large}) and constrained semantic parsing. However, the sequential Monte Carlo algorithms we describe here would be difficult to implement efficiently using \texttt{GenGPT3.jl}. One challenge is that the OpenAI API is stateless, and so ``one-token-at-a-time'' generation and conditioning would require prohibitively many calls (each with growing numbers of context tokens). 

\textbf{Steering language models with programs.} \citet{beurer2022prompting} recently coined the term \textit{language model programming}, to refer to the use of specialized programming languages to guide the behavior of LLMs. Several such programming languages have 
since been introduced, including their SQL-inspired \textit{language model query language} (LMQL), \citet{microsoft2023guidance}'s Guidance language, and \citet{normal2023outlines}'s {Outlines} language. 
All three of these provide high-level DSLs that make chaining multiple calls to LLMs more ergonomic, and {Outlines} and LMQL also expose some features for generation subject to hard constraints. However, they do not support sampling the posterior \textit{given} these constraints, only beam search (in the case of LMQL) and greedy decoding, using token masking to enforce the constraints. Furthermore, the constraint DSLs supported by these languages are limited, and cannot, for example, encode our prompt intersection task. That said, they support many features that would be useful to include in future versions of \texttt{LLaMPPL} (or higher-level DSLs built on it): a unified frontend to a variety of Transformer backends, automated computation of token masks for enforcing common constraints, and high-level syntax for chaining prompts together that does not require explicit token-by-token processing logic.



\textbf{Controlled generation and probabilistic inference in language models.} Many recent papers have proposed methods for more controlled text generation from language models, either through the lens of optimization \citep{kumar2021controlled} or probabilistic inference \citep{kumar2022constrained}. Approaches applied during fine-tuning include direct preference optimization \citep{rafailov2023direct}, reinforcement learning from human feedback \citep{ouyang2022training}, and generation with distributional control \citep{khalifa2021distributional}, all of which can be viewed as forms of variational Bayesian inference due to their use of KL-divergence penalties \citep{korbak2022rl}. Finetuning methods have the benefit of avoiding increased runtime during decoding, but they typically cannot handle hard constraints, motivating the use of controlled generation at decoding time. 

Among decoding-time approaches, many are focused on optimization, either through beam search \citep{meister2020if} and heuristic search \citep{lu2021neurologic,zhang2023planning}, or through gradient-based optimization in embedding space \citep{dathathri2019plug,kumar2021controlled}. Other approaches focus on sampling from a constrained or modified distribution \citep{zhang2023tractable}, including naive rejection sampling \citep{poesia2022synchromesh}, but also more sophisticated Markov Chain Monte Carlo (MCMC) samplers \citep{miao2019cgmh,hie2022high} that make use of specialized proposal distributions \citep{zhang2020language} or the gradients of continuous embeddings \citep{qin2022cold,kumar2022constrained}. However, a downside of MCMC methods is that they require potentially many iterations before producing a sample from the desired distribution, limiting their speed and usefulness. In contrast, SMC steering maintains the autoregressive nature of both regular decoding and beam search while still allowing constraints to be applied. As such, our method achieves the same overhead as beam search, while continuing to sample from the desired distribution instead of optimizing. This enables SMC steering to generate a diversity of constrained completions without the need for additional machinery \citep{vijayakumar2016diverse}.

Researchers have also proposed using the posterior distributions of probabilistic generative models defined in part using Transformers for tasks beyond constrained generation, e.g. semantic segmentation and household navigation, where the general world knowledge learned by the LLM is used to inform priors~\citep{li2023lampp}. Probabilistic programming tools like $\texttt{LLaMPPL}$, which support building models that use LLMs and performing efficient inference in them, could help make such approaches more accessible and scalable.




\bibliographystyle{plainnat}
\bibliography{refs}

\end{document}